%% file: main.tex
\documentclass{article}

    \usepackage[preprint]{neurips_2025}

\usepackage[utf8]{inputenc} 
\usepackage[T1]{fontenc}    
\usepackage{hyperref}       
\usepackage{url}            
\usepackage{booktabs}       
\usepackage{amsfonts}       
\usepackage{nicefrac}       
\usepackage{microtype}      
\usepackage{xcolor}         

\usepackage{amsmath,amssymb} 
\usepackage{graphicx} 
\usepackage{bbding}
\usepackage{makecell}
\usepackage{bbm}
\usepackage{multirow} 
\title{DSG-World: Learning a 3D Gaussian World Model from Dual State Videos}

\author{%
Wenhao~Hu\textsuperscript{1,2}, Xuexiang~Wen\textsuperscript{2}, Xi~Li\textsuperscript{1}, Gaoang~Wang\textsuperscript{1,2}\\
\textsuperscript{1}College of Computer Science and Technology, Zhejiang University\\
\textsuperscript{2}ZJU-UIUC Institute, Zhejiang University\\
}

\begin{document}

\maketitle

\input{tex/0_abs}
\input{tex/1_intro}

\input{tex/2_related}

\input{tex/3_method}

\input{tex/4_dataset}

\input{tex/5_exp}
\input{tex/6_conclusion}

\bibliography{ref}
\bibliographystyle{plain}

\end{document}

%% file: tex/0_abs.tex
\begin{abstract}
Building an efficient and physically consistent world model from limited observations is a long-standing challenge in vision and robotics. 
Many existing world modeling pipelines are based on implicit generative models, which are hard to train and often lack 3D or physical consistency. 
On the other hand, explicit 3D methods built from a single state often require multi-stage processing—such as segmentation, background completion, and inpainting—due to occlusions. To address this, we leverage two perturbed observations of the same scene under different object configurations. These \textit{dual states} offer complementary visibility, alleviating occlusion issues during state transitions and enabling more stable and complete reconstruction. In this paper, we present \textbf{DSG-World}, a novel end-to-end framework that explicitly constructs a 3D \textbf{G}aussian \textbf{World} model from \textbf{D}ual \textbf{S}tate observations. Our approach builds dual segmentation-aware Gaussian fields and enforces bidirectional photometric and semantic consistency. We further introduce a pseudo-intermediate state for symmetric alignment, and design collaborative co-pruning and co-pasting strategies to refine geometric completeness. DSG-World enables efficient real-to-simulation transfer purely in the explicit Gaussian representation space, supporting high-fidelity rendering and object-level scene manipulation without relying on dense observations or multi-stage pipelines. Extensive experiments demonstrate strong generalization to novel views and scene states, highlighting the effectiveness of our approach for real-world 3D reconstruction and simulation.
\end{abstract}

%% file: tex/1_intro.tex
\section{Introduction}
\label{intro}

Constructing reliable world models from sparse observations remains a fundamental challenge in vision and robotics~\cite{zhu20243d, wang2024nerf, panglearning,mendonca2023structured,bar2024navigation,zhen20243d,bear2023unifying}. While recent methods often leverage implicit generative frameworks~\cite{yang2023learning, wu2024ivideogpt, zhu2025unified, wen2024vidman}, these approaches typically suffer from training instability and lack explicit 3D structure or physical grounding, limiting their applicability in tasks requiring interaction, control, or simulation. In contrast, recent advances in 3D Gaussian Splatting~\cite{kerbl20233d} enable explicit scene reconstruction by representing geometry and appearance with compact Gaussian primitives. 

Some methods, such as Gaussian Grouping~\cite{ye2023gaussian}, attempt real-to-simulation (\textit{real2sim}) transformation by combining instance-level segmentation with inpainting-based refinement. While partially effective, these multi-stage pipelines face several challenges. Inaccurate 2D segmentation—especially near object boundaries or under occlusion—leads to imprecise masks that propagate into 3D, causing misclassified Gaussians and visible artifacts. These issues often require additional post-processing, increasing system complexity. Moreover, inpainting struggles to recover fine background details in cluttered or high-frequency regions, resulting in unrealistic or blurry outputs. Such limitations undermine the fidelity and robustness of the real2sim process and hinder downstream tasks requiring accurate scene understanding.
\begin{figure}[t]
    \centering
    \includegraphics[width=\linewidth]{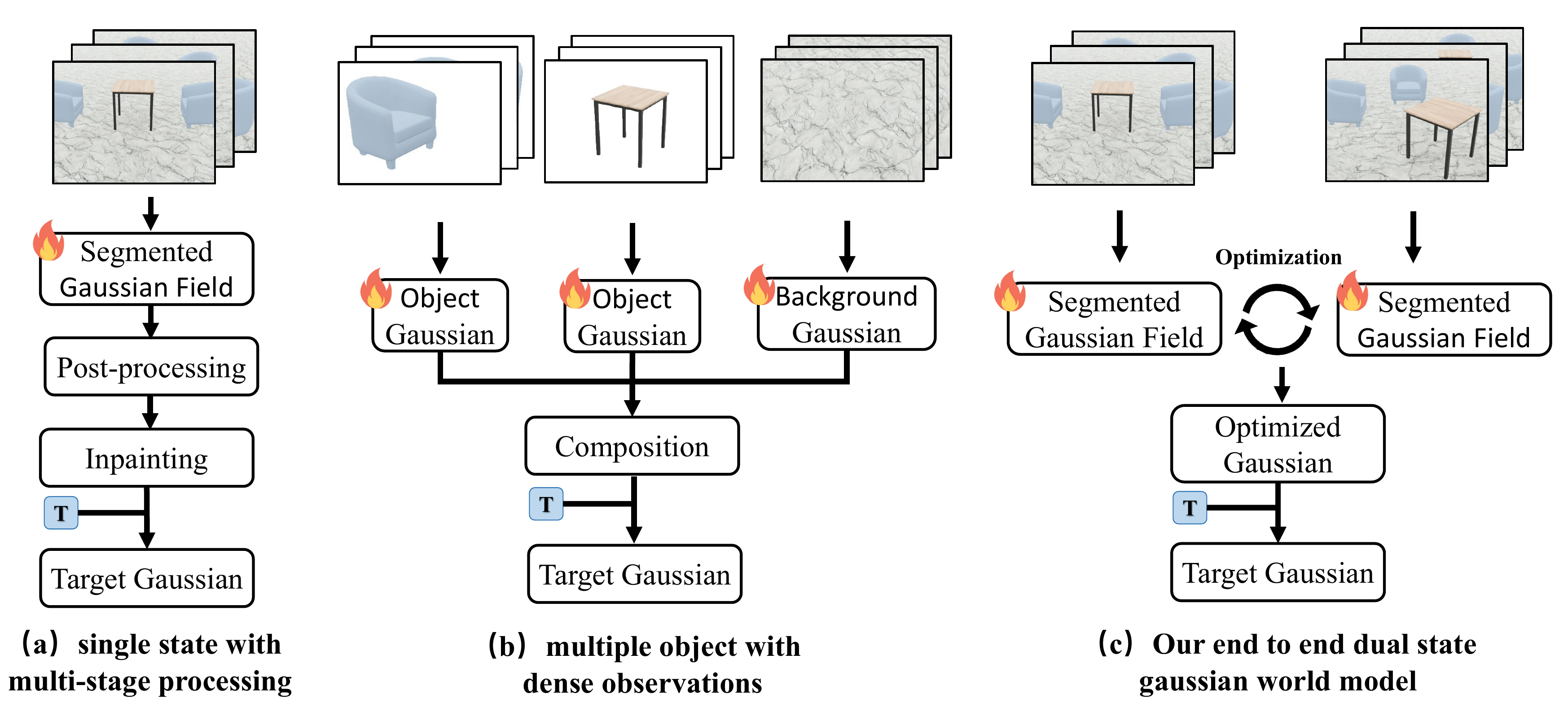}
  \caption{Comparison of different paradigms for constructing 3D Gaussian world models. 
(a) Traditional single-state pipelines rely on multi-stage post-processing and inpainting, which may introduce accumulated artifacts. 
(b) Object-centric approaches require dense multi-view observations of all components, followed by explicit composition. 
(c) Our proposed end-to-end dual-state Gaussian world model jointly optimizes two complementary segmented Gaussian fields via cross-state supervision, enabling accurate and compact target Gaussian reconstruction without requiring inpainting or dense observations.}
    \label{fig:teaser}
\end{figure}

In parallel, recent research in real2sim transfer and world modeling~\cite{barcellona2024dream,yu2025real2render2real,yang2025novel,lou2024robo,han2025re,zhu2025vr} has begun exploring the integration of Gaussian Splatting into interactive and physically grounded simulation frameworks. Methods such as RoboGSim~\cite{li2024robogsim} and SplatSim~\cite{qureshi2024splatsim} leverage Gaussian representations to construct photorealistic virtual environments from real-world observations. However, these approaches heavily rely on dense multi-view object captures to build high-fidelity scene representations, which limits their scalability in practical deployment scenarios. Moreover, their primary focus remains on enhancing appearance realism or facilitating policy transfer, rather than explicitly modeling object-level state transitions.

We argue that remains unresolved: \textbf{how can we efficiently build an accurate and consistent world model of a scene from a minimal number of real observations, while avoiding multi-stage processing and multi-object scanning?} 

To tackle this, we leverage two perturbed observations of the same scene under different object configurations. These \textit{dual states} provide complementary supervision—revealing background regions that may be occluded in one state but visible in the other. Moreover, the relative motion between states helps delineate clearer object boundaries, enhancing the spatial precision of 3D Gaussian segmentation.

Motivated by these insights, we propose \textbf{DSG-World}, a unified framework that constructs a \textbf{D}ual \textbf{S}tate \textbf{G}aussian \textbf{World} model from only two static observations. Our method builds dual segmentation-aware Gaussian fields and applies bidirectional supervision to jointly optimize geometric and semantic consistency. To further enhance alignment, we introduce a pseudo-intermediate state as a shared reference space for fusion and simulation. Finally, we design collaborative co-pruning and co-pasting strategies to improve geometric completeness. Unlike traditional pipelines that rely on multi-stage heuristics or inpainting, DSG-World directly operates in the explicit Gaussian space, enabling efficient and high-fidelity simulation under arbitrary scene configurations.

In summary, our work makes the following contributions: 
\begin{itemize}
\item We present a novel dual-state modeling strategy that constructs a complete and consistent 3D Gaussian field from two perturbed scene states, simplifying state simulation and significantly improving the accuracy and robustness of world model transitions.
\item We enforce mutual consistency by transforming each Gaussian field into the configuration of the other and supervising both photometric and semantic outputs in both directions.
\item  We construct a virtual intermediate Gaussian field via geometric constraints, serving as a symmetric reference to facilitate alignment and comparison across the two observed states.
\item We support object-level novel scene simulation by applying rigid transformations directly in the Gaussian representation space, without relying on latent representations.
\end{itemize}

%% file: tex/2_related.tex
\section{Related Works}

\subsection{3D Gaussian Segmentation} 
Recent methods have extended Gaussian Splatting to perform scene segmentation~\cite{zhu2025rethinking,hu2025pointmap,hu2024semantic}.
GaussianEditor~\cite{chen2024gaussianeditor} projects 2D segmentation masks back onto 3D Gaussians via inverse rendering. Language-driven approaches like LangSplat~\cite{qin2024langsplat}, LEGaussians~\cite{shi2024language}, and others~\cite{chen2024ovgaussian,qiu2024gls} leverage CLIP features for open-world scene understanding.
Gaussian Grouping~\cite{ye2023gaussian} attaches segmentation features to each Gaussian and aligns multi-view segment IDs using video segmentation methods~\cite{cheng2023tracking}, while Gaga~\cite{lyu2024gaga} further addresses ID inconsistency across views through a 3D-aware memory bank.
FlashSplat~\cite{shen2024flashsplat} introduces a fast, globally optimal linear programming-based approach for 3D Gaussian segmentation.
OpenGaussian~\cite{wu2024opengaussian} and InstanceGaussian~\cite{li2024instancegaussian} use contrastive learning to obtain point-level 3D segmentation.
GaussianCut~\cite{jain2024gaussiancut} models the scene as a graph and employs graph-cut optimization to separate foreground and background Gaussians.
COB-GS~\cite{zhang2025cob} improves segmentation accuracy via boundary-adaptive Gaussian splitting and visual refinement, enabling sharper boundaries while maintaining rendering fidelity.

Obtaining a 3D segmentation alone is insufficient for real2sim simulation, as 2D segmenter biases often lead to inaccurate 3D masks, requiring post-processing and Gaussian inpainting~\cite{liu2024infusion,cao2024mvinpainter, huang20253d} to fill background holes caused by object movement. This multi-stage pipeline is complex and prone to error accumulation. In contrast, our method leverages two complementary scene states to provide mutual visibility, enabling end-to-end reconstruction without explicit inpainting. Object transfers across states help calibrate imperfect segmentations, resulting in a clean and consistent Gaussian world model.

\subsection{Interactive World Modeling and Real2Sim Simulation} 
Some approaches construct implicit generative world models from video data. UniSim\cite{yang2023learning} simulates real-world interactions by predicting visual outcomes conditioned on diverse actions using a unified autoregressive framework over heterogeneous datasets.
iVideoGPT\cite{wu2024ivideogpt} models interactive dynamics by encoding visual observations, actions, and rewards into token sequences for scalable next-token prediction with compressive tokenization. However, these methods lack 3D and physical consistency and are generally difficult to train.

Recent methods aim to build interactive simulators by integrating reconstructed real scenes into physics engines. For instance, RoboGSim~\cite{li2024robogsim} embeds 3D Gaussians into Isaac Sim; SplatSim~\cite{qureshi2024splatsim} replaces meshes with Gaussian splats for photorealistic rendering; PhysGaussian~\cite{xie2024physgaussian} and Spring-Gaus~\cite{zhong2024reconstruction} enable mesh-free physical simulation with Newtonian or elastic models; NeuMA~\cite{cao2024neuma} refines simulation using image-space gradients. However, these approaches typically rely on dense, per-object 3D capture. In contrast, our method is much simpler and more lightweight—requiring only two real videos and known rigid transformations. By explicitly constructing dual-state Gaussian fields, we achieve consistent and efficient Real2Sim conversion without multi-object scanning or complex simulation components.

%% file: tex/3_method.tex
\section{Preliminary and Problem Definition}
\label{pre}
\subsection{Segmented Gaussian Splatting}  
Segmented Gaussian Splatting~\cite{ye2023gaussian} models a scene as a set of 3D Gaussians, each parameterized as $\mathcal{G} = \{\boldsymbol{x}, \boldsymbol{\Sigma}, \boldsymbol{\alpha}, \boldsymbol{c}, \boldsymbol{s}\}$, where $\boldsymbol{x}$ denotes the 3D center position, $\boldsymbol{\Sigma}$ represents the spatial covariance matrix, $\boldsymbol{\alpha}$ is the opacity coefficient, $\boldsymbol{c}$ is the RGB color vector, and $\boldsymbol{s}$ is a learnable feature vector used for segmentation. 

During rendering, each Gaussian is projected onto the 2D image plane using a differentiable $\alpha$-blending mechanism. Both the final pixel color $C$ and segmentation feature $S$ are computed by accumulating Gaussian contributions weighted by their projected opacities $\boldsymbol{\alpha}_i'$:
\begin{align}
C = \sum_{i \in \mathcal{N}} \boldsymbol{c}_i \boldsymbol{\alpha}_i' \prod_{j=1}^{i-1}(1 - \boldsymbol{\alpha}_j'), \quad
S = \sum_{i \in \mathcal{N}} \boldsymbol{s}_i \boldsymbol{\alpha}_i' \prod_{j=1}^{i-1}(1 - \boldsymbol{\alpha}_j')
\label{eq:rendering}
\end{align}

\subsection{Motivation and Core Definition}

We define the world model as a process of \textit{explicit object-level state transition}, where changes in the scene are modeled as discrete transitions between physically grounded states. Specifically, when an object undergoes a rigid transformation and relocates within the scene, it is interpreted as a transition from one explicit state to another. Unlike generative models that implicitly predict future states in a latent space---often resulting in physically inconsistent or ambiguous interpretations---our formulation maintains spatial and physical coherence by directly modeling the geometry and motion of objects in 3D space.

\subsection{Modeling with Dual-State Observations}

Constructing a world model from a single static observation is inherently difficult, as it typically relies on segmentation, post-processing, and inpainting---each prone to introducing errors or inconsistencies. To mitigate these challenges, we propose building the world model from \textbf{two static observations} captured from distinct scene states. By leveraging the \textit{object-level motion} between these states, we jointly refine 3D segmentation and recover occluded background regions. This enables extrapolation to arbitrary intermediate states and facilitates the construction of a consistent 3D Gaussian world model. Formally, the modeling process is defined as:
\begin{align}
(\mathcal{G}_1^*, \mathcal{G}_2^*) = \arg\min_{\mathcal{G}_1, \mathcal{G}_2} \; \mathcal{L}_{\text{joint}}
\label{eq:optimization}
\end{align}
Given the optimized pair $(\mathcal{G}_1^*, \mathcal{G}_2^*)$ and  object-level transformation $T$, we synthesize a new target Gaussian field $\mathcal{G}_t$ through explicit transformation:
\begin{align}
\mathbf{W}_{GS} : \{\mathcal{G}_1^*, \mathcal{G}_2^*\}\xrightarrow{T} \mathcal{G}_t.
\label{eq:state_transfer}
\end{align}

\subsection{Gaussian State Transfer}

Let $\mathcal{I}_1$ and $\mathcal{I}_2$ denote the image sets captured from two observed scene states, with corresponding segmentation masks $\mathcal{S}_1$ and $\mathcal{S}_2$. Given a transformation $\mathbf{T}$ that describes the motion of foreground objects between state 1 and state 2, our goal is to reconstruct the 3D geometry of both states using Gaussian primitives and to enable simulation and interpolation across different scene configurations.

We first construct two Gaussian fields, $\mathcal{G}_1$ and $\mathcal{G}_2$, from $(\mathcal{I}_1, \mathcal{S}_1)$ and $(\mathcal{I}_2, \mathcal{S}_2)$, respectively. Each field encodes geometry and appearance information of the scene in its corresponding state. The transformation between the fields is then formulated as:
\begin{align}
\mathcal{G}_2 = \mathbf{T}_{1\to2}(\mathcal{G}_1), \quad \mathcal{G}_1 = \mathbf{T}_{2\to1}(\mathcal{G}_2)
\end{align}

\subsection{Object-Aware Transformation Modeling}

To model scene-level transformations, the operator $\mathbf{T}$ is defined as an object-aware function that applies per-Gaussian rigid transformations based on semantic identity. Let the Gaussian field be decomposed into foreground and background subsets:
\begin{align}
\mathcal{G} = \mathcal{G}_\text{fg} \cup \mathcal{G}_\text{bg}, \quad \mathcal{G}_\text{fg} = \bigcup_{o=1}^{O} \mathcal{G}_\text{fg}^{(o)}
\end{align}
where each foreground object $o$ is associated with a rigid transformation $\mathbf{T}^{(o)}$. For any Gaussian $\boldsymbol{g}_i \in \mathcal{G}$, let $o_i$ denote the object it belongs to. Then, $\mathbf{T}$ is applied as:
\begin{align}
\mathbf{T}(\boldsymbol{g}_i) =
\begin{cases}
\mathbf{T}^{(o_i)} \cdot \boldsymbol{g}_i, & \text{if } \boldsymbol{g}_i \in \mathcal{G}_\text{fg} \\
\boldsymbol{g}_i, & \text{if } \boldsymbol{g}_i \in \mathcal{G}_\text{bg}
\end{cases}
\end{align}
This formulation ensures spatially consistent transformation and geometric fidelity of object-level motion while preserving the static background.

\section{Method}
\label{method}
\begin{figure}[ht]
    \centering
    \includegraphics[width=\linewidth]{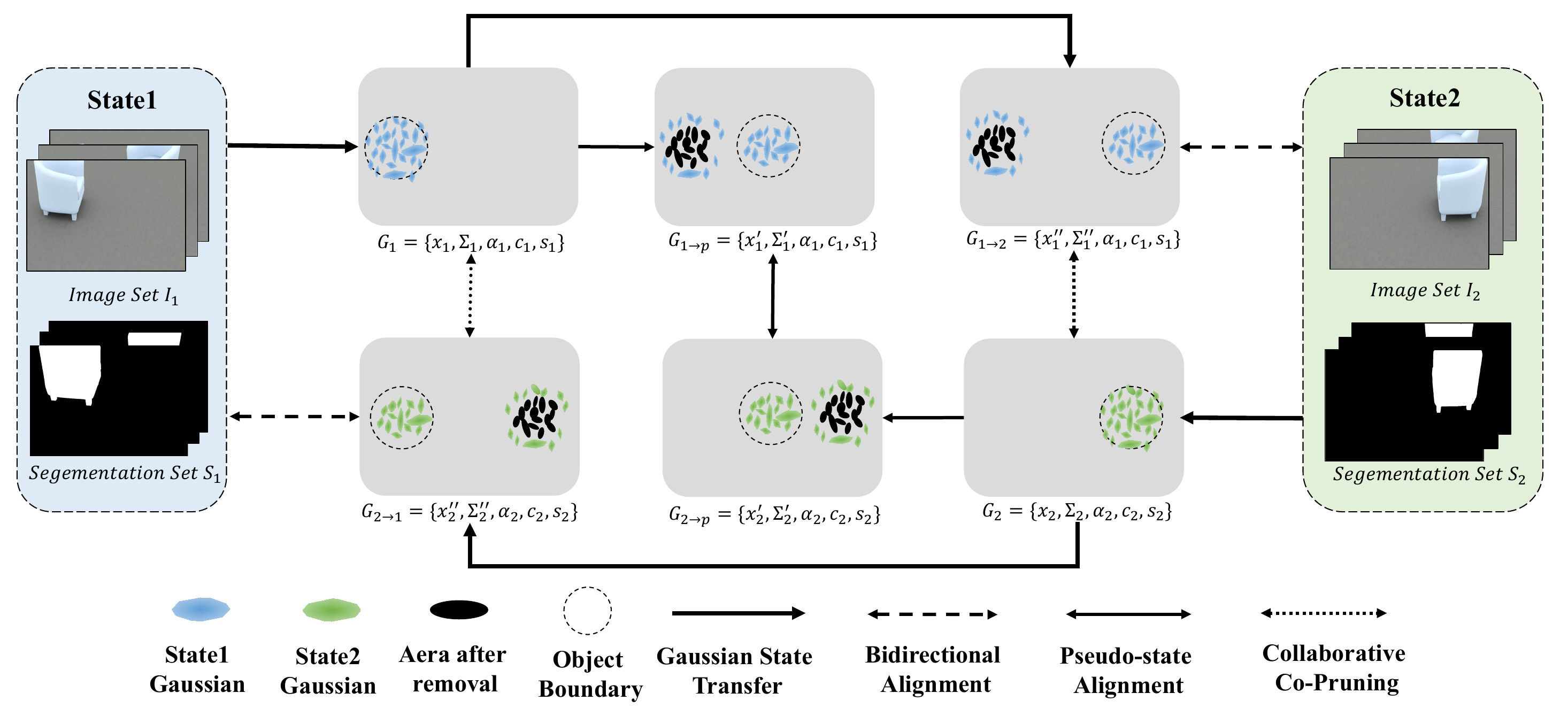}
    \caption{Overview of our dual-state Gaussian optimization pipeline. Given two complementary scene observations from different states (State1 and State2), we construct two segmented Gaussian fields, $G_1$ and $G_2$. During state transfer (e.g., $G_1 \rightarrow G_{1 \rightarrow 2}$), object motion is applied to the segmented region. However, Gaussians near object boundaries (dashed circles) often remain static due to incomplete or inaccurate boundary segmentation during training. This leads to residual inconsistencies across states, particularly at the edges of moving objects. To enable realistic and artifact-free image under arbitrary configurations in simulation, we introduce a bidirectional alignment mechanism and pseudo-state supervision ($G_{1 \rightarrow p}$ and $G_{2 \rightarrow p}$) to regularize the transferred Gaussians and promote coherent boundary alignment. A co-pruning strategy is further employed to refine consistency and remove redundant or misaligned components.}

    \label{fig:pipeline}
\end{figure}

\subsection{Bidirectional Alignment}
\label{sec:Mapping}
To ensure geometric and semantic consistency across different scene states, we enforce that the rendered outputs from transformed Gaussian fields align with the ground-truth observations in the corresponding target states. As mentioned before, we apply transformation $\mathbf{T}_{1\to2}$ to $\mathcal{G}_1$ and transformation $\mathbf{T}_{2\to1}$ to $\mathcal{G}_2$. For any viewpoint $v$, the transformed Gaussian fields are rendered into RGB images and segmentation masks, which are then compared with the corresponding ground-truth observations $(\mathcal{I}_1^v, \mathcal{S}_1^v)$ and $(\mathcal{I}_2^v, \mathcal{S}_2^v)$ from the original states. The total alignment loss combines photometric and segmentation consistency, defined as:
\begin{align}
\mathcal{L}_{\text{align}}(\mathcal{G}_1,\mathcal{G}_2,\theta) = &\left\| R\left( \mathbf{T}_{1\to2}(\mathcal{G}_1), v \right) - \mathcal{I}_2^v \right\|_1 
+ \left\| R\left( \mathbf{T}_{2\to1}(\mathcal{G}_2), v \right) - \mathcal{I}_1^v \right\|_1 \nonumber \\
&+ \mathrm{CE}\left( f_\theta\left(\mathcal{M}\left( \mathbf{T}_{1\to2}(\mathcal{G}_1), v \right)\right), \mathcal{S}_2^v \right) 
+ \mathrm{CE}\left( f_\theta\left(\mathcal{M}\left( \mathbf{T}_{2\to1}(\mathcal{G}_2), v \right)\right), \mathcal{S}_1^v \right)
\label{eq:align_loss}
\end{align}
where $R(\cdot, v)$ and $\mathcal{M}(\cdot, v)$ denote the rendered RGB image and segmentation feature, respectively, from viewpoint $v$, as defined in Equation~\ref{eq:rendering}. The segmentation output is obtained via a shared classifier $f_\theta$, which is jointly applied to both $\mathcal{G}_1$ and $\mathcal{G}_2$. Specifically, $f_\theta$ consists of a linear layer that projects each identity embedding to a $(K+1)$-dimensional space, where $K$ is the number of instance masks in the 3D scene~\cite{ye2023gaussian}. The cross-entropy loss $\mathrm{CE}(\cdot, \cdot)$ measures the semantic alignment between predicted and ground-truth masks. 

This bidirectional consistency encourages each transformed Gaussian field to accurately reconstruct the scene content of the opposite state, thereby reinforcing object-level correspondence and enhancing alignment across different scene configurations.

\subsection{Pseudo-state Guided Alignment}
\label{sec:Pseudo}

To enhance the generalizability of the world model across diverse scene configurations, we introduce a pseudo-state $\mathcal{G}_p$ that serves as an intermediate reference for supervision. This pseudo-state is constructed by applying geometric constraints, such as collision and boundary regularization, to synthesize a virtual configuration between the two observed states. Unlike the original states, the pseudo-state is not tied to any specific observation but provides a common frame that facilitates consistent alignment between $\mathcal{G}_1$ and $\mathcal{G}_2$.

We compute transformation matrices $\mathbf{T}_{1\to{p}}$ and $\mathbf{T}_{2\to{p}}$ to transfer the original fields $\mathcal{G}_1$ and $\mathcal{G}_2$ into $\mathcal{G}_p$. By transforming both fields into this shared pseudo-state, we enable direct comparison and alignment of their rendered outputs. Specifically, we render the transformed fields from the same viewpoint $v$ and enforce photometric and semantic consistency between them. The corresponding loss is defined as:
\begin{align}
\mathcal{L}_{\text{pseudo}}(\mathcal{G}_1,\mathcal{G}_2,\theta) = 
&\left\| R\left( \mathbf{T}_{1\to{p}}(\mathcal{G}_1), v \right) 
- R\left( \mathbf{T}_{2\to{p}}(\mathcal{G}_2), v \right) \right\|_1 \nonumber \\
&+ \mathrm{CE}\left( f_\theta\left(\mathcal{M}\left( \mathbf{T}_{1\to{p}}(\mathcal{G}_1), v \right)\right), 
f_\theta\left(\mathcal{M}\left( \mathbf{T}_{2\to{p}}(\mathcal{G}_2), v \right)\right) \right)
\label{eq:pseudo_loss}
\end{align}
By leveraging a dynamically constructed pseudo-state as an adaptive supervision signal, the model can better reconcile differences between the two input states and generalize more effectively to unseen or intermediate scene configurations.

\subsection{Collaborative Co-Pruning}
\label{sec:Pruning}
To suppress residual artifacts introduced by imperfect segmentation during cross-state Gaussian transfer, we propose a co-pruning mechanism that filters out spatially inconsistent Gaussians by leveraging geometric agreement between the two states. When a Gaussian field is transferred from one state to another, unmatched or misaligned points may remain due to occlusion, noise, or over-segmentation. Our strategy prunes these outliers by checking whether transferred Gaussians can be reliably explained by the geometry of the target field.

For each transformed Gaussian $\boldsymbol{g}_i \in \mathbf{T}_{1\to2}(\mathcal{G}_1)$, we identify its nearest neighbor $\boldsymbol{g}_j \in \mathcal{G}_2$ using Euclidean distance. A Gaussian is marked for pruning if the spatial deviation between $\boldsymbol{g}_i$ and $\boldsymbol{g}_j$ exceeds a predefined threshold $\tau$. The binary pruning indicator $m_i$ is computed as:
\begin{align}
m_i =  \mathbbm{1}\left( \left\| \boldsymbol{x}_i - \boldsymbol{x}_j \right\|_2 > \tau \right),
\end{align}
where $\boldsymbol{x}_i$ and $\boldsymbol{x}_j$ are the 3D centers of $\boldsymbol{g}_i$ and $\boldsymbol{g}_j$, and $\mathbbm{1}(\cdot)$ denotes the indicator function. Gaussians with $m_i = 1$ are discarded as unreliable or redundant. A symmetric process is applied in the opposite direction, using $\mathcal{G}_2$ transformed to the frame of $\mathcal{G}_1$ to prune outliers in $\mathcal{G}_2$, resulting in a collaborative co-pruning scheme.

\subsection{Training Objective}
The overall training objective combines three loss terms:
\begin{align}
\mathcal{L}_{joint}(\mathcal{G}_1,\mathcal{G}_2,\theta) = \mathcal{L}_{\text{r}}(\mathcal{G}_1,\theta) + 
\mathcal{L}_{\text{r}}(\mathcal{G}_2,\theta)+ \lambda_a \mathcal{L}_{\text{align}}(\mathcal{G}_1,\mathcal{G}_2,\theta) + \lambda_p \mathcal{L}_{\text{pseudo}}(\mathcal{G}_1,\mathcal{G}_2,\theta),
\end{align}
where $\mathcal{L}_{\text{r}}$ denotes the same reconstruction loss adopted from Gaussian Grouping~\cite{ye2023gaussian} (detailed in the appendix), $\mathcal{L}_{\text{align}}$ enforces bidirectional rendering consistency, and $\mathcal{L}_{\text{pseudo}}$ introduces regularization through pseudo-state supervision. The weights $\lambda_a$ and $\lambda_p$ are used to balance the contributions of each term.

%% file: tex/4_dataset.tex
\section{Dataset}
\label{dataset}

To support dual-state scene modeling, we construct both synthetic and real-world datasets. The synthetic dataset is generated in Blender~\cite{blender}, where $N$ textured objects from BlenderKit~\cite{blenderkit} are placed in a static background. A second state is created by altering object poses, ensuring no ground region is occluded in both states to preserve complementary visibility. Real-world data is captured similarly using handheld RGB cameras. The dataset includes 7 synthetic and 5 real scenes.

For evaluation, we create a test state for each scene by randomly repositioning objects. We render images from test-view cameras and compute PSNR and SSIM against the ground truth, measuring simulation fidelity under novel configurations. More implementation and dataset details are provided in the appendix.

%% file: tex/5_exp.tex
\section{Experiment}
\label{exp}
\subsection{Experimental Setup}
\paragraph{Implementation details}
During training, we first optimize the segmented Gaussians using only $\mathcal{L}_{\text{recon}}$ for 10,000 epochs, then jointly train with $\mathcal{L}_{\text{align}}$ and $\mathcal{L}_{\text{pseudo}}$ to refine the dual Gaussian field for another 10,000 epochs.
The output classification linear layer has 16 input channels and 256 output channels. 
The pruning threshold parameter $\tau$ is set to 0.5. In training, we set $\lambda_a=1.0$  and $\lambda_=1.0$. We use the Adam optimizer for both gaussians and linear layer, with a learning rate of 0.0025 for segmentation feature and 0.0005 for linear layer. All datasts are trained for 20K iterations on a single NVIDIA 4090 GPU.

\paragraph{Baselines}
We compare our method with representative Gaussian Splatting-based simulation frameworks, applying necessary adaptations for fair evaluation. Existing pipelines often involve multi-stage processing, including segmentation, background completion, inpainting, and fine-tuning. We include segmentation methods based on inverse rendering (GaussianEditor~\cite{chen2024gaussianeditor}), feature-based segmentation (Gaussian Grouping~\cite{ye2023gaussian}), and graph optimization (GaussianCut~\cite{jain2024gaussiancut}); Gaussian Grouping* denotes the variant with convex hull filtering. To enhance simulation quality, we apply kNN-based feature propagation from Gaussian Grouping and LaMa~\cite{suvorov2022resolution} inpainting for hole filling, and extend the GraphCut setup with depth-aware Gaussian completion. We also include Decoupled Gaussian~\cite{wang2025decoupledgaussian}, which segments objects using Gaussian Grouping, then performs remeshing and LaMa-based refinement to complete the scene.

\subsection{Novel State Simulation}
\begin{figure}[!ht]
    \centering
    \includegraphics[width=1\linewidth]{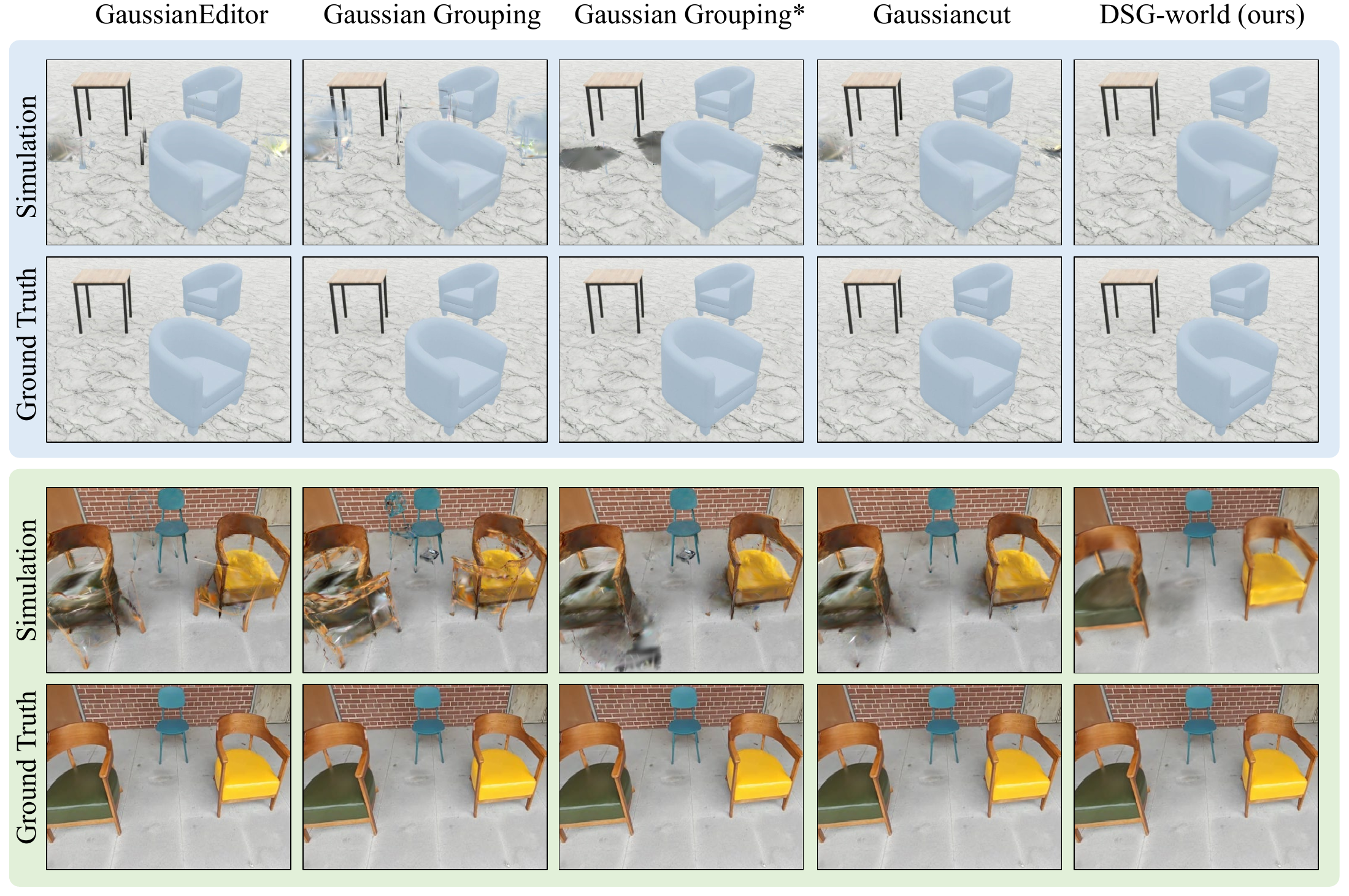}
    \caption{Qualitative comparison of simulated scene reconstruction under different segmentation-based Gaussian pipelines. We evaluate on both synthetic (top) and real-world (bottom) scenes. While existing methods (e.g., GaussianEditor, Gaussian Grouping, and Gaussiancut) struggle with object mixing, boundary artifacts, or background corruption, our method (DSG-world) achieves significantly more accurate and complete simulation results, closely matching the ground-truth scene configuration.}

    \label{novel}
\end{figure}

\begin{figure}[!ht]
    \centering
    \includegraphics[width=1\linewidth]{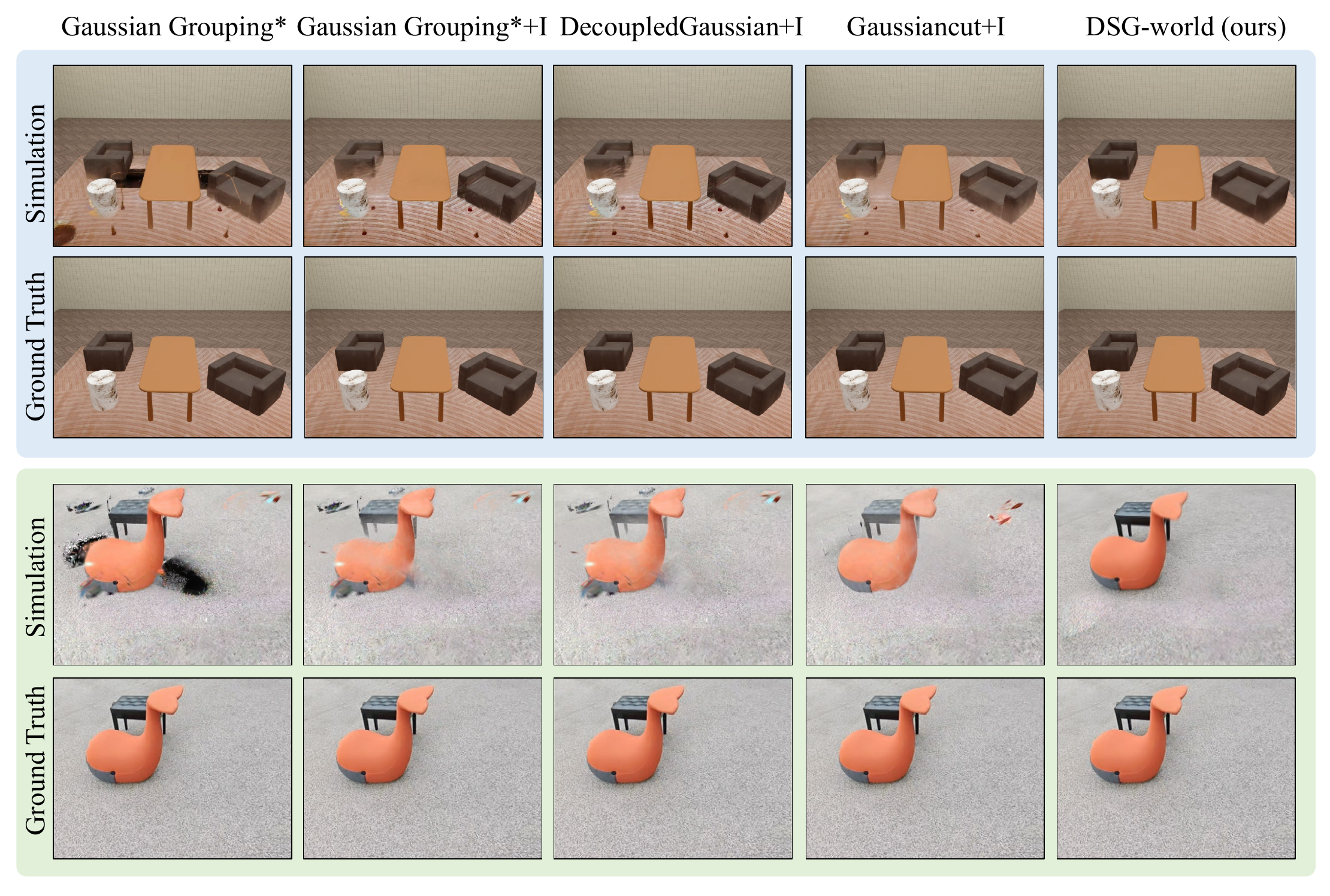}
    \caption{Visual comparison of synthetic (top) and real-world (bottom) scenes simulation across different segmentation and inpainting pipelines. We evaluate multiple baseline methods including Gaussian Grouping*, Gaussian Grouping*+I (with inpainting), DecoupledGaussian+I, and Gaussiancut+I, against our proposed DSG-world on both synthetic (top) and real (bottom) scenes.}

    \label{novel_i}
\end{figure}
The qualitative results on both virtual and real datasets are shown in Figure~\ref{novel}. GaussianEditor (inverse rendering based) fails to segment object boundaries precisely, causing edge artifacts that require heavy post-processing for simulation. Gaussian Grouping (segmentation-feature based) often miss inner-object features, leading to floating Gaussians after motion. Graussiancut (graph-based) performs best among baselines, though slight boundary artifacts persist.

For fair comparison, we adopt a unified post-processing pipeline for methods that exhibit significant scene gaps after object movement, as shown in Fig.~\ref{novel_i}. This pipeline includes background Gaussian completion, 2D inpainting, and Gaussian finetuning. The KNN-based completion in Gaussian Grouping often leads to occlusion of existing objects. Gaussiancut with depth-based completion and LaMa inpainting achieves the most visually coherent result. DecoupledGaussian, which builds upon Gaussian Grouping, provides remeshing performance comparable to depth completion but still introduces floating artifacts. In contrast, our method achieves the highest PSNR and SSIM for novel-state simulation across both datasets, while remaining end-to-end and avoiding complex multi-stage post-processing.

\begin{table}[ht]
 \caption{Quantitative comparison of simulation quality across different 3D Gaussian segmentation pipelines on the synthetic and real-world datasets. Methods are categorized into segmentation-only, segmentation with inpainting (Seg.+Inpainting), and our proposed dual-state framework (DSG-world)}
  \label{tab:2d_data}
  \centering
  \begin{tabular}{l|l|c|cc|cc}
    \toprule
    \textbf{Type} & \textbf{Model} & \textbf{Completion} & \multicolumn{2}{c|}{\textbf{Sim}} & \multicolumn{2}{c}{\textbf{Real}} \\
    & & & PSNR & SSIM & PSNR & SSIM \\
    \midrule
    \multirow{4}{*}{Segmentation}
    & GaussianEditor~\cite{chen2024gaussianeditor} & - & 25.82 & 0.929 & 23.25 & 0.801 \\
    & Gaussian Grouping~\cite{ye2023gaussian} & - & 26.22&	0.913&	22.74	&0.777 \\
    & Gaussian Grouping*~\cite{ye2023gaussian} & - & 22.29&	0.901	&22.37&	0.790 \\
    & Gaussiancut~\cite{jain2024gaussiancut} & - & 26.79&	0.941&	23.43&	0.809 \\
    \midrule
    \multirow{3}{*}{Seg.+Inpainting}
    & Gaussian Grouping*~\cite{ye2023gaussian} & knn & 
    29.31	&0.892&	23.28	&0.805\\
    & DecoupledGaussian~\cite{wang2025decoupledgaussian} & remesh & 29.50&	0.891&	24.28&	0.804 \\
    & Graphcut~\cite{jain2024gaussiancut} & depth & 
    30.88&	0.937&	24.40&	0.810\\
    \midrule
    Dual-State & DSG-world (ours) & Co-pasting &
\textbf{38.37} & \textbf{0.974} & \textbf{27.52} & \textbf{0.859} \\
    \bottomrule
  \end{tabular}
\end{table}

As shown in Table~\ref{tab:2d_data}, segmentation-only methods yield lower PSNR and SSIM, while adding inpainting improves performance—particularly on synthetic scenes with large, textured floors that expose noticeable holes after object movement. In contrast, real-world scenes typically involve smaller movable objects and mostly textureless floors, resulting in less visible gaps. Instead of relying on inpainting, we leverage the complementary nature of dual-state Gaussians to directly retrieve ground Gaussians from the alternate state for completion, a process we define as Co-pasting. As the results show, completion-based methods perform similarly overall, while our dual-state approach achieves the highest accuracy on both datasets.

\subsection{Ablation Study}
Table~\ref{tab:ablation} presents the ablation study evaluating the contribution of each component in our framework: Bidirectional Alignment (B), Collaborative Co-Pruning (C), and Pseudo-state Guided Alignment (P). Using only Bidirectional Alignment already provides a strong baseline, achieving a PSNR of 36.57. Introducing Co-Pruning yields a slight improvement in structural quality. This is because Bidirectional Alignment tends to reassign residual Gaussians to have background-like colors or reduced opacity. While Co-Pruning helps eliminate these floaters, its overall impact on PSNR is limited. In contrast, incorporating Pseudo-state Guided Alignment results in a substantial increase in PSNR. This improvement arises from the fact that occlusion ambiguities cannot be fully resolved with only two configurations, additional pseudo-states provide richer supervision across multiple viewpoints, enhancing alignment between the two Gaussian fields and leading to more consistent and photorealistic reconstructions.

\begin{table}[ht]
  \centering
  \renewcommand{\arraystretch}{0.9}
  \begin{minipage}[t]{0.48\textwidth}
    \centering
    \caption{Ablation study of B (Bidirectional Alignment), C (Collaborative Co-Pruning), and P (Pseudo-state Guided Alignment).}
    \label{tab:ablation}
    \begin{tabular}{ccc|cc}
      \toprule
      B & C & P & PSNR~$\uparrow$ & SSIM~$\uparrow$ \\
      \midrule
      \CheckmarkBold & - & - & 36.57 & 0.974 \\
      \CheckmarkBold & \CheckmarkBold & - & 36.96 & 0.977 \\
      \CheckmarkBold & \CheckmarkBold & \CheckmarkBold & \textbf{38.37} & \textbf{0.974} \\
      \bottomrule
    \end{tabular}
  \end{minipage}
  \hfill
  \begin{minipage}[t]{0.48\textwidth}
    \centering
    \caption{PSNR and SSIM of $\mathcal{G}_1$ and $\mathcal{G}_2$  in Sim1 and Sim2 scene, with and without pseudo-state supervision.}
    \label{tab:psnr_ssim}
    \begin{tabular}{c|cc|cc}
      \toprule
       & \multicolumn{2}{c|}{\textbf{Sim1}} & \multicolumn{2}{c}{\textbf{Sim2}} \\
      & PSNR & SSIM & PSNR & SSIM \\
      \midrule
      $\mathcal{G}_1$ & 37.04 & 0.979 & 37.51 & 0.9774 \\
      $\mathcal{G}_2$ & 37.16 & 0.977 & 36.14 & 0.9747 \\
      \midrule
      $\mathcal{G}_1^*$ & 39.26 & 0.9844 & 37.59 & 0.9776 \\
      $\mathcal{G}_2^*$ & 39.26 & 0.9843 & 37.50 & 0.9774 \\
      \bottomrule
    \end{tabular}
  \end{minipage}
\end{table}

\subsection{Dual Guassian Convergence}

We investigate the convergence behavior of two Gaussian fields trained from different scene states. Without Pseudo-state Guided Alignment, we denote the fields as $\mathcal{G}_1$ and $\mathcal{G}_2$, and with Pseudo-state Guided Alignment, as $\mathcal{G}_1^*$ and $\mathcal{G}_2^*$. In the absence of Pseudo-state Guided Alignment, PSNR and SSIM differ significantly when evaluated in the target state. These discrepancies stem from occlusions and viewpoint differences that lead to misalignments between the two fields. Even with Bidirectional Alignment, such inconsistencies persist, indicating incomplete convergence.

By incorporating Pseudo-state Guided Alignment, we enforce consistency across object compositions in both fields, allowing them to observe complementary content and provide mutual supervision. This promotes convergence toward a shared and coherent optimized representation. Empirically, Gaussian fields trained from either state yield nearly identical PSNR and SSIM when evaluated under the same configuration, demonstrating effective alignment and mutual consistency.

%% file: tex/6_conclusion.tex
\section{Limitations and Broader impacts}
\label{limit}
Despite its effectiveness, DSG-world has several limitations. To ensure end-to-end reconstruction, dual-state observations are designed to be complementary, avoiding occlusion overlap. However, if object perturbations are too minor, some regions may remain unobserved in both states, hindering accurate reconstruction. Additionally, our model does not handle lighting variations, causing static shadows even when objects move, which affects realism. Incorporating relighting into the framework could further enhance simulation fidelity in future work.

\section{Conclusion}
\label{conclusion}
We present DSG-World, an end-to-end framework for 3D world modeling from dual-state observations. By leveraging complementary visibility, our method mitigates occlusion issues and removes the need for complex post-processing. Through bidirectional consistency and pseudo-state alignment, it generates high-quality Gaussian fields for accurate rendering and simulation. Extensive results demonstrate its effectiveness and scalability for real-to-simulation tasks in vision and robotics.